\documentclass{article}

\usepackage{arxiv}

\usepackage[utf8]{inputenc} 
\usepackage[T1]{fontenc}    
\usepackage{hyperref}       
\usepackage{url}            
\usepackage{booktabs}       
\usepackage{amsfonts}       
\usepackage{nicefrac}       
\usepackage{microtype}      
\usepackage{graphicx}
\usepackage{natbib}
\usepackage{doi}
\usepackage{subfig}

\title{\textbf{Effect Of Personalized Calibration On Gaze Estimation Using Deep-Learning}}

\author{
 Nairit Bandyopadhyay\hspace{10mm}
 Sébastien Riou\hspace{10mm} 
 Didier Schwab\hspace{10mm}\\
 \\Univ. Grenoble Alpes, CNRS, Grenoble INP, LIG, 38000 Grenoble, France\\
 \\\texttt{Nairit.Bandyopadhyay@grenoble-inp.org}\\
 \\\texttt{\{sebastien.riou, didier.schwab\}@univ-grenoble-alpes.fr}
}

\date{}

\hypersetup{
pdftitle={Effect Of Personalized Calibration On Gaze Estimation Using Deep-Learning},
pdfauthor={Nairit~Bandyopadhyay, Sébastien~Riou,Didier~Schwab},
pdfkeywords={Gaze tracking, Deep Learning, Personal Calibration},
}

\begin{document}
\maketitle

\begin{abstract}
	With the increase in computation power and the development of new state-of-the-art deep learning algorithms, appearance-based gaze estimation is becoming more and more popular. It is believed to work well with curated laboratory data sets, however it faces several challenges when deployed in real world scenario. One such challenge is to estimate the gaze of a person about which the Deep Learning model trained for gaze estimation has no knowledge about. To analyse the performance in such scenarios we have tried to simulate a calibration mechanism. In this work we use the MPIIGaze data set. We trained a multi modal convolutional neural network and analysed its performance with and without calibration and this evaluation provides clear insights on how calibration improved the performance of the Deep Learning model in estimating gaze in the wild.
\end{abstract}


\section{Introduction}
Increase in interactions between human and computing devices is leading to the popularity of research in gaze-tracking. Specific domains like Human Computer Interaction and Computer Vision are researching on the various applications of appearance based gaze estimation or tracking. \citep{article1}. 
Machine learning methods can be used to train gaze-estimators, provided we have lots of reliable data, and head pose-independent training data \citep{inproceedings1,inproceedings2}. These methods can bring appearance-based methods into a certain state where these methods do not require any user or device-specific training. The quantity of user data is sufficient to provide information to the gaze-estimators. Gaze tracking using monocular cameras in mobile phones, laptops and interactive displays are cost effective because of their widespread availability. While appearance-based gaze estimation performs well with machine learning-based methods, new techniques are still developed. These new techniques are evaluated on curated data sets collected under controlled laboratory conditions \citep{inproceedings3}. As a result there is lack of availability of different eye shapes as well as we have to make assumptions that the head pose is accurate. This often results in problems in \textit{object recognition} \citep{inproceedings4} and \textit{object detection} \citep{inproceedings5}.\\[0.1cm]
In our study we have used MPIIGaze data set. \textit{The data set and annotations are publicly available online} \citep{inproceedings3}.\\[0.1cm]
Here in this article we are going to see how introduction of small amounts of test data in the training set can significantly increase the accuracy of the model. We are calling this approach as a personal calibration that is done in gaze estimation software - when a user starts using it for the first time.\\[0.1cm]
We have divided our work in two parts. First, we train a CNN model for appearance-based gaze estimation. \textit{The data set we are using is one order of magnitude larger than existing data sets and has more variation with respect to illumination and appearance} \citep{inproceedings3}. Second, we perform experiments on the model with and without calibration to see the effect in performance of the model.

\section{Dataset and its features}
We used \textit{MPIIGaze dataset that contains a total of 213,659 images from 15 participants. The number of images of each participant varied from 34,745 to 1,498. The data set contains larger variability in illumination and larger appearance variations} \citep{inproceedings3}. All these data were collected from the laptop because it can be used for daily recordings and are an important platform of eye-tracking devices \citep{incollection1}.

\section{Method}
\begin{figure}[h]
\includegraphics[width=\textwidth]{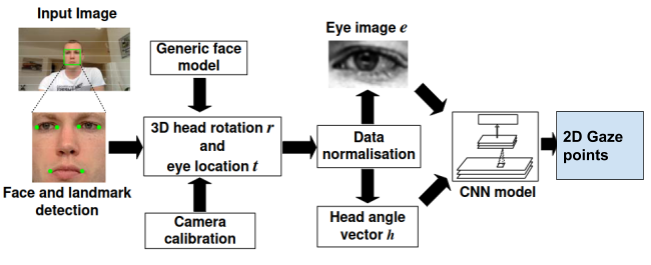}
\caption{Overview of our method for appearance-based gaze estimation using multi modal convolutional neural networks \citep{inproceedings3}.}
\label{figure:Figure 1}
\end{figure}

Figure \ref{figure:Figure 1} provides a high level view of gaze estimation task using multi modal convolutional neural networks (CNN). A monocular camera is used to capture the image of the face. Then the face detection and facial landmarks detection methods are used to locate landmarks in the image \citep{inproceedings3}. A 3D facial shape model is used to estimate 3D poses of the detected faces and apply the space normalisation technique to crop and warp the head pose and eye images to the normalised training space \citep{inproceedings1,inproceedings3}. Then we train a convolutional neural network to learn the mapping from the head poses and eye images to 2D gaze-points in the camera coordinate system.

\subsection{Preprocessing : Face Alignment and 3D Head Pose Estimation}
Before using the images to train the model, the images present in the MPIIGaze data set needs to be processed. \textit{The user’s face is detected in the image using Li et al.’s SURF cascade method} \citep{li2013learning,inproceedings3}. Afterwards, \textit{Baltrušaitis et al.’s constrained local mode framework to detect facial landmarks} \citep{baltruvsaitis2014continuous,inproceedings3}. The 3D positions of six facial landmarks (eye and mouth corners, cf.
Figure \ref{figure:Figure 1}) constitutes the facial model. The head coordinate system is defined according
to the triangle connecting three midpoints of the eyes and
mouth. \textit{EPnP algorithm is used to fit the model by estimating the initial solution and further refining the pose via non-linear optimisation} \citep{lepetit2009epnp,inproceedings3}. 3D head rotation \textit{\textbf{r}} is defined as
the \textit{rotation from the head coordinate system to the camera coordinate system}, and the eye position \textit{\textbf{t}} is defined as the \textit{midpoint of eye corners for each eye} \citep{inproceedings3}.

\begin{figure}[h]
\includegraphics[width=\textwidth]{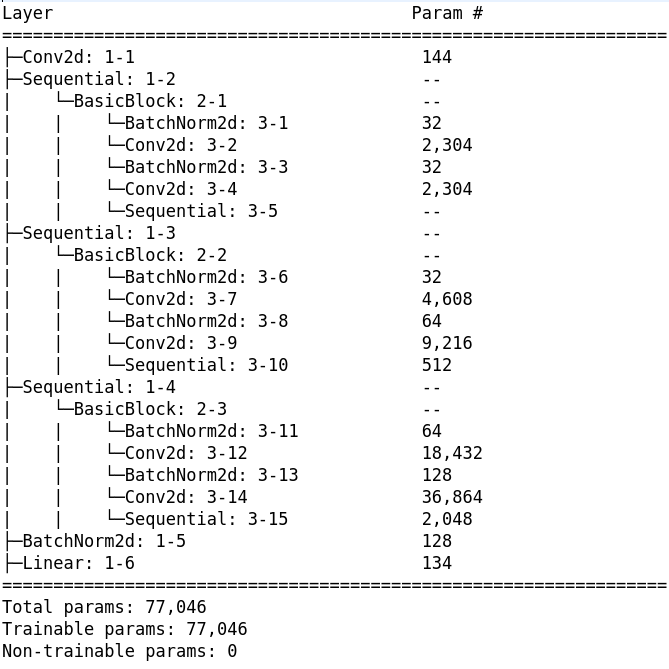}
\caption{ResNet Model Summary}
\label{figure:Figure 2}
\end{figure}

\subsection{Normalizing the Preprocessed Data}
The normalisation is done by two camera matrix transformation operations: scaling and rotation of camera \citep{inproceedings3}. The x axes of the camera coordinate space and that of head coordinate space is made parallel by pointing the camera to one of the facial landmarks - the midpoint of eye corners. \textit{After that, the eye images were cropped at a fixed resolution \textit{\textbf{W}} × \textit{\textbf{H}} with a fixed focal length \textit{\textbf{f}} in the normalised camera space, and histogram-equalised (to improve the contrast of the image) to form the input eye image. This results in a set of fixed-resolution eye images \textit{\textbf{e}} and 2D head angle vectors \textit{\textbf{h}}} \citep{inproceedings3}. In order to reduce the effect of different lighting conditions, so that no extra noise gets inside the CNN model, eye images \textit{\textbf{e}} are histogram-equalised after the normalisation process \citep{inproceedings3}. \textit{The setting for camera distance
\textit{\textbf{d}}, focal length \textit{\textbf{f}} and the resolution \textit{\textbf{W}} ×\textit{\textbf{H}} remains same} \citep{inproceedings1}.

\subsection{Using Multimodal CNN To Learn The Mapping}
\textit{The CNN learns the mapping from the input features (2D head angle h and eye image e) to 2D points g in the screen normalised space} \citep{inproceedings1,inproceedings3}.\\[0.1cm]
Our model uses a pre-activated ResNet like architecture that consists of one convolutional layer containing 3 stages, followed by batch normalisation and a final connected layer. There is a linear regression layer on top of the final connected layer to predict gaze positions \textit{\textbf{g}}. The head pose information is introduced in our CNN model by concatenating \textit{\textbf{h}} with the output of the fully connected layer. Our input for the network are the \textit{grey-scale eye images \textit{\textbf{e}} with a fixed size of 60 ×36 pixels} \citep{inproceedings3}. The number of parameters for the convolutional layer is 144. For more details on the architecture please refer to Figure \ref{figure:Figure 2} The output of the network is a 2D gaze position g that consists of x and y coordinates (normalised) of the 2D screen. Our loss function is the sum of the individual losses that measure the euclidean distance between the predicted \textit{\textbf{g}} and the actual \textit{\textbf{g}}. Thus with the neural network we are solving an optimisation problem of minimising this loss function\\[0.1cm]
The normalized data set is available online for public use. Hence we did not have to perform the techniques in Section 3.1 and Section 3.2. We mainly worked on Section 3.3.

\begin{figure}[h]
\includegraphics[width=\textwidth]{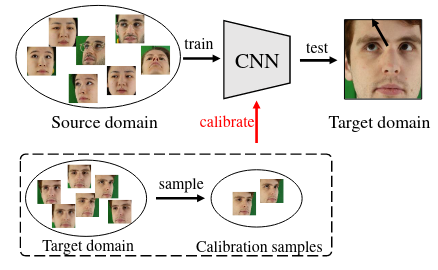}
\caption{Higher level representation of personal calibration method for Deep Learning technique \citep{cheng2021appearance}}
\label{figure:Figure 3}
\end{figure}

\section{Experiments}
Here we will be discussing about the person-independent gaze estimation task with and without calibration to validate the effectiveness of the proposed CNN based gaze estimation approach with personalized calibration. There are various ways of personal calibration, like fine-tuning the model in target domain \citep{krafkaeyetracking}. Some researchers have added some target person specific features for gaze tracking which are learned during fine tuning \citep{linden2019learning}. We will be retraining the model with the target feature and observe if there are any changes in the performance of the model, with and without personalized calibration. Please refer to this Figure \ref{figure:Figure 3} to understand the pipeline of personalized calibration.

\subsection{Equipment Used}
All the training were done using GeForce GTX 1650 Max-Q. The average time to train 1 epoch having 12 steps is around 20 seconds at around 90\% GPU utilisation. 

\subsection{Methodology}
We conduct the experiments using the MPIIGaze dataset only. \textit{We choose 1,500 left eye samples and 1,500 right eye samples from each person so that we can take into account the sample number bias in the data set. Since one participant has only 1,448 images, we randomly over sampled the data to get 3,000. So for each person, we now have 3000 images} \citep{inproceedings3}. We split these 3000 images into 10 partitions, each containing 10\% i.e 300 images for calibration. We used leave-one-out evaluation strategy. This means, if we want to evaluate the performance of a model on the data of person \textit{X}, then we train a model using data of the remaining persons and some small percentage of data of person \textit{X}. Here we have used 10\% of the test data for calibration. Then we test the resulting model \textit{A} with the remaining 90\% data of the person \textit{X}. Again, we also have to train a model \textit{B} that is without calibration and test with the same 90\% test data that we used to test model \textit{A}. This will now give us the difference in estimation error with and without calibration. For each person, we perform this experiment 10 times (because we have 10 partitions) so as to get the mean estimation error and see which partition provides best calibration for each person.\\[0.5cm]
We trained every model for 40 epochs because we found that around 40 epochs, the training loss and validation loss converges and also the loss function is minimum at that time.

\begin{figure}[h]
\includegraphics[width=\textwidth]{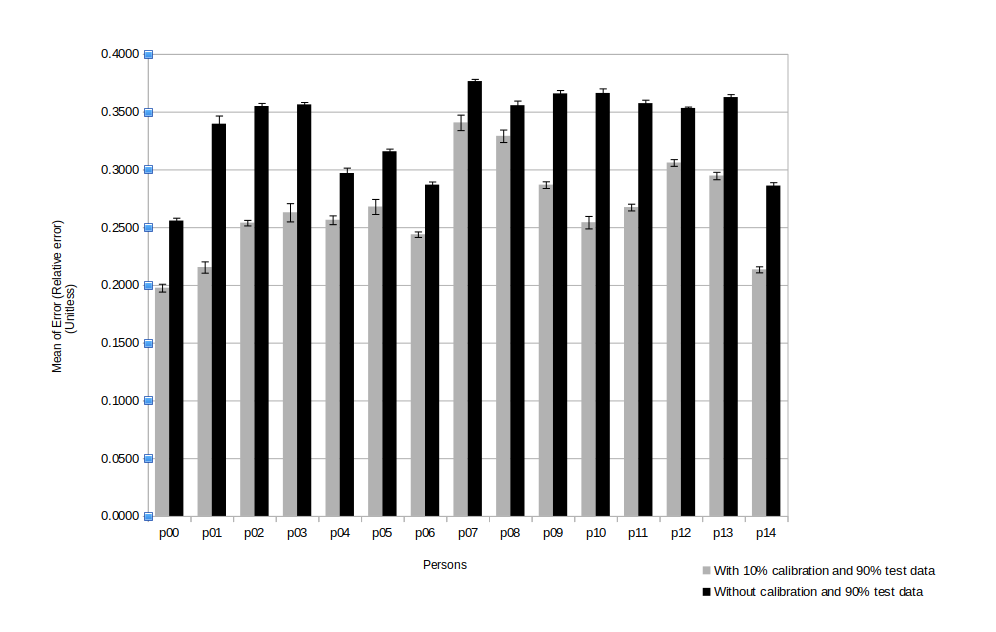}
\caption{Leave-one-person-out evaluation on MPIIGaze. Mean estimation errors are plotted for each person for the proposed method of personalised calibration. Error bars indicate standard deviations when the experiments are conducted with and without calibration.}
\label{figure:Figure 4}
\end{figure}

\section{Results}
As described in Section 5.1, the experiments were conducted. We had to conduct 150 experiments and in each experiment we had to create two models - with and without calibration. For each experiment we obtain two error values. It is important to note that the error value here is unit less because it is a relative error. Due to differences in screen size for each participant, the ground truth pixel coordinates values were normalized by dividing the x and y pixel coordinate by width and height of the screen in pixels respectively so that the pixel coordinate values remain between 0 and 1. Hence the error value also remains in between 0 and 1. For training and testing of model, such normalization was necessary, or else the model was not getting trained due to some out of range errors. However for real time setup where the model infers the gaze position on the screen, in that case the x,y coordinate values can be multiplied with dimensions of the screen so as to get the exact pixel position. The results obtained showed an interesting observation that calibration significantly reduces the mean error compared to without calibration. This is because the model now has some prior information about the person (See Figure \ref{figure:Figure 4}).\\[0.5cm]
Let's look at the results of p00. The mean error is 0.25 before calibration. It means this 0.25 is the difference in the ground truth value and the inferred value of the pixel coordinate. As mentioned before, it has no unit because the pixel coordinates were normalised. After calibration, it is now 0.20. There has been a significant improvement in case of p00 where the error decreased by almost 20\%.
\section{Limitations}
One of the major limitation is the process of head-pose estimation in real time. Since we used the normalized dataset of MPIIGaze, we did not have to perform the head-pose estimation. However if someone decides to create a new dataset of their own, then they have to perform this head-pose estimation.
Sometimes the detection of facial landmarks and obtaining eye-images are also challenging because of lighting conditions and presence of accessories like when a person wears glasses. We did not have to do this as we used the clean normalized MPIIGaze dataset.
\begin{figure}%
    \centering
    \subfloat[\centering Facial landmark detection]{{\includegraphics[width=7.5cm]{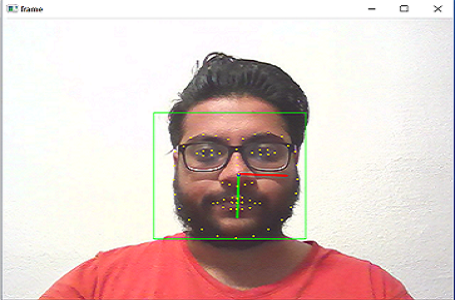} }}%
    \qquad
    \subfloat[\centering Predicted x,y coordinates - for left eye and right eye separately along with the head position]{{\includegraphics[width=7.5cm]{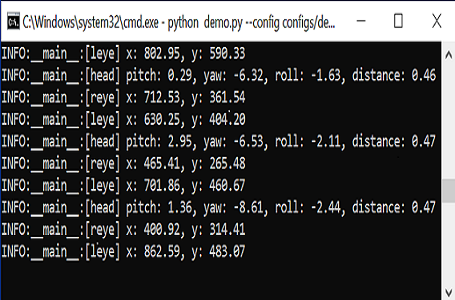} }}%
    \caption{Real Time 2D Gaze Points Estimation}%
    \label{fig:example}%
\end{figure}
\section{Future Work}
By using only 300 extra images (10\% of the test data) we have seen significant decrease in estimation errors. So our next work will be to create a real-time software with this calibration method. Before the software starts gaze-tracking, it will capture images of the face of the person using webcam. Most modern day webcams can shoot 30 times in one second. So we can capture around 450 images in 15 seconds by which we can calibrate a pre-trained model. After that we can start using that software to get a predicted 2-D position on the screen.\\[0.1cm]
We have already started creating such software which is still in development stage. As of now, it does not have the calibration feature and also we have not tested it rigorously. It can do the tasks of facial landmark detection and predicting the x,y coordinates of the estimated eye-gaze on a 2D screen. Please refer to Figure \ref{fig:example} which shows our software working in real time.\\[0.1cm]
Apart from this, we can also try using other models like UNet etc. to see if it improves the performance of learning by the model. Thus, there are multiple scope of future works based on this work.

\section{Conclusion}
Appearance-based gaze estimation methods have so far been evaluated exclusively under controlled laboratory conditions. In this work, we present an extensive study on appearance-based gaze estimation along with a calibration technique. We used a data set that contains images which has wide variations. Our CNN-based estimation model significantly shows improvement in performance when calibration is performed. This work provides a critical insight on addressing the performance challenges of Deep Learning models in daily-life gaze interaction.

\bibliographystyle{achemso}
\bibliography{template}  






\end{document}